\documentclass{article}
\usepackage{spconf,graphicx}
\usepackage{mathtools}
\usepackage{cite}
\usepackage{tipa}
\usepackage{amsmath,amssymb}
\DeclareMathOperator{\E}{\mathbb{E}}

\title{Deep learning the EEG manifold for phonological categorization from active thoughts}

\twoauthors
 {Pramit Saha, Sidney Fels}
	{Human Communication Technologies Lab,\\ University of British Columbia}
 {Muhammad Abdul-Mageed }
	{Natural Language Processing Lab,\\ University of British Columbia}
\begin{document}
%
\maketitle
\begin{abstract}
Speech-related Brain Computer Interfaces (BCI) aim primarily at finding an alternative vocal communication pathway for people with speaking disabilities. As a step towards full decoding of imagined speech from active thoughts, we present a BCI system for subject-independent classification of phonological categories exploiting a novel deep learning based hierarchical feature extraction scheme. To better capture the complex representation of high-dimensional electroencephalography (EEG) data, we compute the joint variability of EEG electrodes into a channel cross-covariance matrix. We then extract the spatio-temporal information encoded within the matrix using a mixed deep neural network strategy. Our model framework is composed of a convolutional neural network (CNN), a long-short term network (LSTM), and a deep autoencoder. We train the individual networks hierarchically, feeding their combined outputs in a final gradient boosting classification step. Our best models achieve an average accuracy of 77.9\% across five different binary classification tasks, providing a significant 22.5\% improvement over previous methods. As we also show visually, our work demonstrates that the speech imagery EEG possesses significant discriminative information about the intended articulatory movements responsible for natural speech synthesis.
\end{abstract}

\section{Introduction}
Decoding intended speech or motor activity from brain signals is one of the major research areas in Brain Computer Interface (BCI) systems~\cite{herff2016automatic,d2009toward}. In particular, speech-related BCI technologies attempt to provide effective vocal communication strategies for controlling external devices through speech commands interpreted from brain signals~\cite{ghane2015silent}. Not only do they provide neuro-prosthetic help for people with speaking disabilities and neuro-muscular disorders like locked-in-syndrome, nasopharyngeal cancer, and amytotropic lateral sclerosis (ALS), but also equip people with a better medium to communicate and express thoughts, thereby improving the quality of rehabilitation and clinical neurology ~\cite{netsell1982speech,martin2016word}. Such devices also have applications in entertainment, preventive treatments, personal communication, games, etc. Furthermore, BCI technologies can be utilized in silent communication, as in noisy environments, or situations where any sort of audio-visual communication is infeasible. 
\begin{figure}[t]
\centering
\includegraphics[width=11cm,height=4.5cm,keepaspectratio]{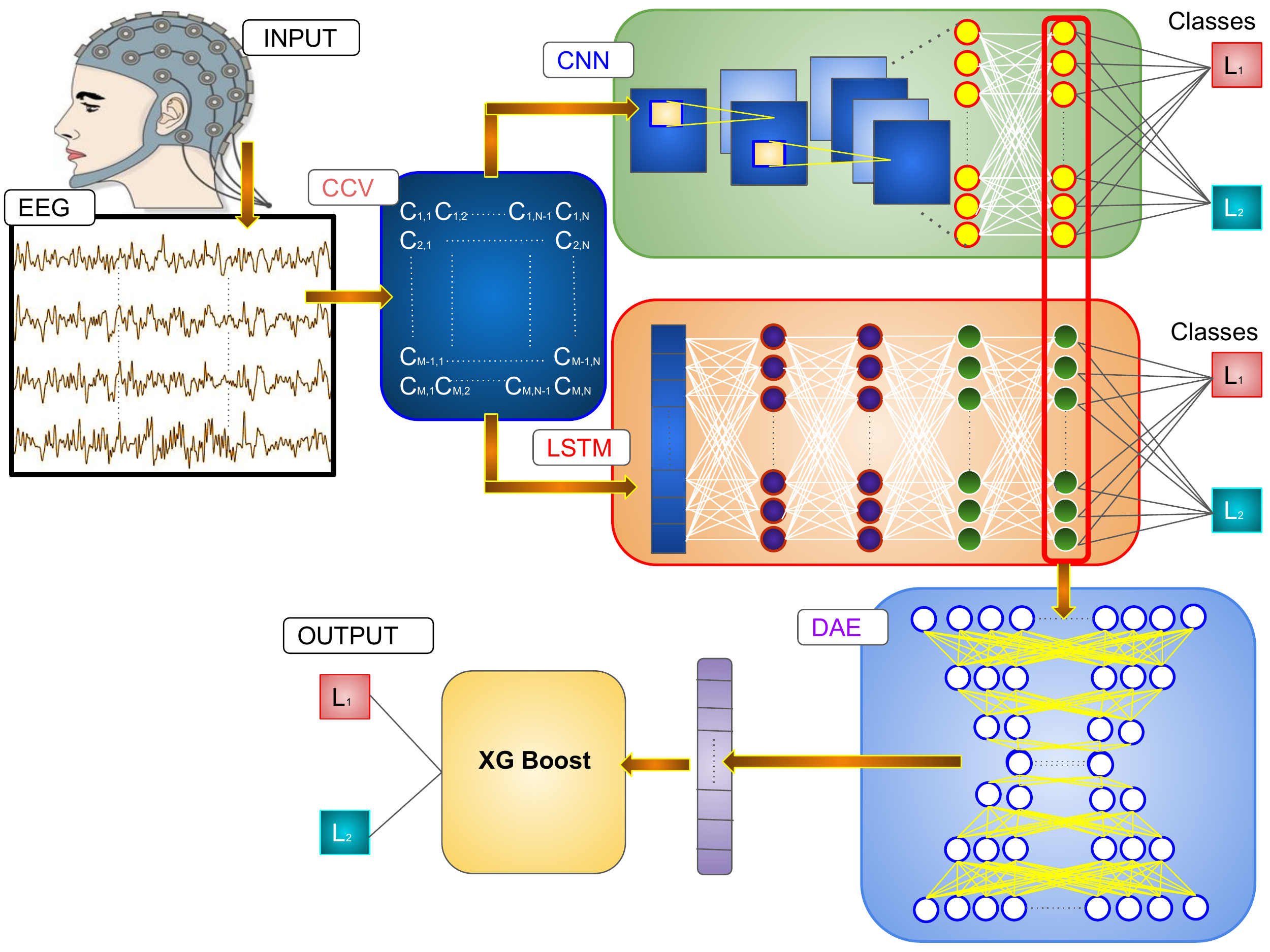}
\caption{Overview of the proposed approach}
\label{fig:framework}
\end{figure}

Among the various brain activity-monitoring modalities in BCI, electroencephalography (EEG)~\cite{pfurtscheller2008eeg, guger1999prosthetic} has demonstrated promising potential to differentiate between various brain activities through measurement of related electric fields. EEG is non-invasive, portable, low cost, and provides satisfactory temporal resolution. This makes EEG suitable to realize BCI systems. EEG data, however, is challenging: these data are high dimensional, have poor SNR, and suffer from low spatial resolution and a multitude of artifacts. For these reasons, it is not particularly obvious how to decode the desired information from raw EEG signals.

Although the area of BCI based speech intent recognition has received increasing attention among the research community in the past few years, most research has focused on classification of individual speech categories in terms of discrete vowels, phonemes and words \cite{dasalla2009single,dasalla2009spatial,idrees2016vowel,deng2010eeg, kim2014eeg,brigham2010imagined,mohanchandra2016communication,wang2013analysis,gonzalez2017sonification}. This includes categorization of imagined EEG signal into binary vowel categories like \textit{/a/}, \textit{/u/} and rest \cite{dasalla2009single,dasalla2009spatial,idrees2016vowel}; binary syllable classes like \textit{/ba/} and \textit{/ku/}~\cite{d2009toward,deng2010eeg,kim2014eeg,brigham2010imagined}; a handful of control words like \textit{'up'}, \textit{'down'}, \textit{'left'}, \textit{'right'} and \textit{'select'} \cite{gonzalez2017sonification} or others like \textit{'water'}, \textit{'help'}, \textit{'thanks'}, \textit{'food'}, \textit{'stop'} ~\cite{mohanchandra2016communication}, Chinese characters~\cite{wang2013analysis}, etc. Such works mostly involve traditional signal processing or manual feature handcrafting along with linear classifiers (e.g., SVMs). In our recent work\cite{saha2018hierarchical}, we introduced deep learning models for classification of vowels and words that achieved 23.45\% improvement of accuracy over the baseline.

Production of articulatory speech is an extremely complicated process, thereby rendering understanding of the discriminative EEG manifold corresponding to imagined speech highly challenging. As a result, most of the existing approaches failed to achieve satisfactory accuracy on decoding speech tokens from the speech imagery EEG data. Perhaps, for these reasons, very little work has been devoted to relating the brain signals to the underlying articulation. The few exceptions include \cite{zhao2015classifying,sun2016neural}. In \cite{zhao2015classifying}, Zhao et al. used manually handcrafted features from EEG data, combined with speech audio and facial features to achieve classification of the phonological categories varying based on the articulatory steps. However, the imagined speech classification accuracy based on EEG data alone, as reported in \cite{zhao2015classifying,sun2016neural}, are not satisfactory in terms of accuracy and reliability. We now turn to describing our proposed models. 
\begin{figure}[t]
\centering
\includegraphics[width=7cm,height=3.5cm,keepaspectratio]{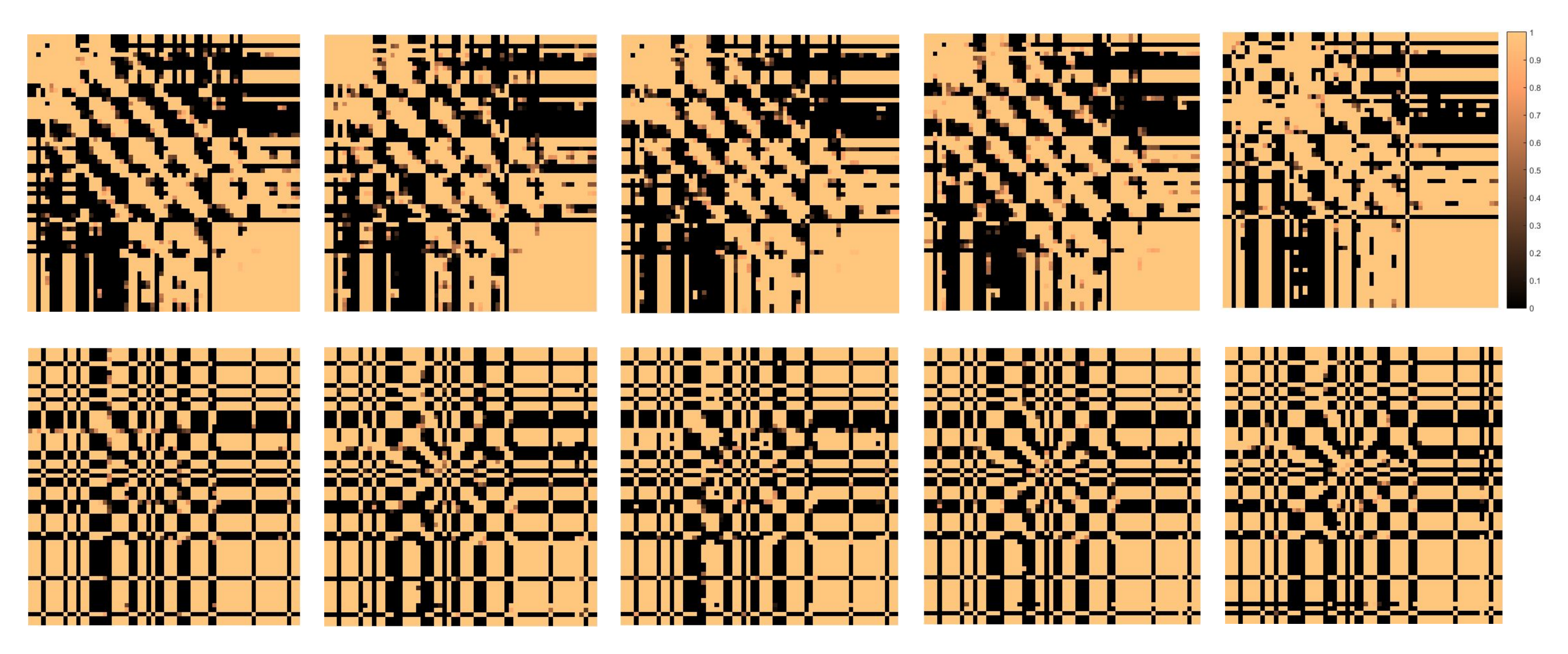}
\caption{Cross covariance Matrices : Rows correspond to two different subjects; Columns (from left to right) correspond to sample examples for bilabial, nasal, vowel, /uw/, and /iy/.}
\label{fig:ccv_matrices}
\end{figure}

\section{Proposed Framework}
Cognitive learning process underlying articulatory speech production involves incorporation of intermediate feedback loops and utilization of past information stored in the form of memory as well as hierarchical combination of several feature extractors. To this end, we develop our mixed neural network architecture composed of three supervised and a single unsupervised learning step, discussed in the next subsections and shown in Fig. \ref{fig:framework}. We formulate the problem of categorizing EEG data based on speech imagery as a non-linear mapping~ $\hat{\textit{\textbf{f}}}$ of a multivariate time-series input sequence ${\textbf{X}^{c}_t}$ to fixed output $\textbf{\textit{y}}$, \textit{i.e}, mathematically ~$\hat{\textbf{\textit{f}}}$~:~ $\textbf{X}^{c}_t~ \longrightarrow~\textbf{\textit{y}}$, where \textit{c} and \textit{t} denote the EEG channels and time instants respectively.

\subsection{Preprocessing step}
We follow similar pre-processing steps on raw EEG data as reported in \cite{zhao2015classifying} (ocular artifact removal using blind source separation, bandpass filtering and subtracting mean value from each channel) except that we do not perform Laplacian filtering step since such high-pass filtering may decrease information content from the signals in the selected bandwidth.

\subsection{Joint variability of electrodes}
Multichannel EEG data is high dimensional multivariate time series data whose dimensionality depends on the number of electrodes. It is a major hurdle to optimally encode information from these EEG data into lower dimensional space. In fact, our investigation based on a development set (as we explain later) showed that well-known deep neural networks (e.g., fully connected networks such as convolutional neural networks, recurrent neural networks and autoencoders) fail to individually learn such complex feature representations from single-trial EEG data. Besides, we found that instead of using the raw multi-channel high-dimensional EEG requiring large training times and resource requirements, it is advantageous to first reduce its dimensionality by capturing the information transfer among the electrodes.  
Instead of the conventional approach of selecting a handful of channels as \cite{zhao2015classifying,sun2016neural}, we address this by computing the channel cross-covariance, resulting in positive, semi-definite matrices encoding the connectivity of the electrodes. We define channel cross-covariance (CCV) between any two electrodes $c_1$ and $c_2$ as:
$Cov({X}^{c_{1}}_t,{X}^{c_{2}}_{t+\tau})=\displaystyle \E[X^{c_{1}}(t)-\mu_{X^{c_{1}}}(t)][X^{c_{2}}(t+\tau)-\mu_{X^{c_{2}}}(t+\tau)]$.
Next, we reject the channels which have significantly lower cross-covariance than auto-covariance values (where auto-covariance implies CCV on same electrode).
We found this measure to be essential as the higher cognitive processes underlying speech planning and synthesis involve frequent information exchange between different parts of the brain. Hence, such matrices often contain more discriminative features and hidden information than mere raw signals. This is essentially different than our previous work~\cite{saha2018hierarchical} where we extract per-channel 1-D covariance information and feed it to the networks. We present our sample 2-D EEG cross-covariance matrices (of two individuals) in Fig. \ref{fig:ccv_matrices}. 

\subsection{CNN \& LSTM}
In order to decode spatial connections between the electrodes from the channel covariance matrix, we use a CNN ~\cite{lecun1989generalization}, in particular a four-layered 2D CNN stacking two convolutional and two fully connected hidden layers. The $k^{th}$ feature map at a given CNN layer with input $x$, weight matrix $W^{k}$ and bias $b_k$ is obtained as: $h^{k}=ReLU(W^{k}*x+b_k)$. At this first level of hierarchy, the network is trained with the corresponding labels as target outputs, optimizing a cross-entropy cost function. In parallel, we apply a four-layered recurrent neural network on the channel covariance matrices to explore the hidden temporal features of the electrodes. Namely, we exploit an LSTM ~\cite{hochreiter1997long} consisting of two fully connected hidden layers, stacked with two LSTM layers and trained in a similar manner as CNN. 

\begin{table}[]
\scriptsize 
 \caption{Selected parameter sets}
\label{hyp}
\begin{tabular}{*{5}{p{1.75cm}}}
\hline\hline
\textbf{Parameters}&\textbf{CNN}  & \textbf{LSTM}& \textbf{DAE}\\ \hline\hline
Batch size    & 64   & 64    & 64   \\ \hline
Epochs    &50     & 50  & 200   \\ \hline
Total layers & 6  & 6 & 7 \\ \hline 
Hidden layers' details    & Conv:32,64 masks:3x3 Dense: 64,128  &  LSTM: 128,256 Dense: 512,1024 &  512,128,32 (Encoder) 
32,128,512 (Decoder)\\ \hline
Activations &  ReLU, last-layer : softmax & all ReLU, last-layer : softmax & ReLU, ReLU, sigm, sigm, ReLU, tanh \\ \hline
Dropout & .25,~~ .50 & .25,~~ .50 & .25,~.25,~ .25
\\ \hline
Optimizer & Adam & Adam & Adam
\\ \hline
Loss & Binary cross entropy & Binary cross entropy & Mean Sq Error
\\ \hline
l-rate & .001 & .001 & .001
\\ \hline\hline
\end{tabular}
\end{table}
\subsection{Deep autoencoder for spatio-temporal information}
As we found the individually-trained parallel networks (CNN and LSTM) to be useful (see Table \ref{test_result}), we suspected the combination of these two networks could provide a more powerful discriminative spatial and temporal representation of the data than each independent network. As such, we concatenate the last fully-connected layer from the CNN with its counterpart in the LSTM to compose a single feature vector based on these two penultimate layers. Ultimately, this forms a joint spatio-temporal encoding of the cross-covariance matrix.   

In order to further reduce the dimensionality of the spatio-temporal encodings and cancel background noise effects\cite{zhang2018converting}, we train an unsupervised deep autoenoder (DAE) on the fused heterogeneous features produced by the combined CNN and LSTM information. The DAE forms our second level of hierarchy, with 3 encoding and 3 decoding layers, and mean squared error (MSE) as the cost function. 

\begin{figure}[]
\centering
\includegraphics[width=7cm,height=5cm,keepaspectratio]{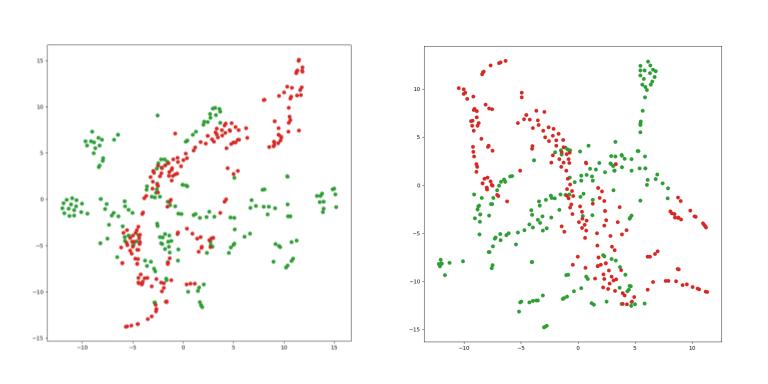}
\caption{tSNE feature visualization for $\pm{nasal}$ (left) and V/C classification (right). Red and green colours indicate the distribution of two different types of features} 
\label{fig:tSNE}
\end{figure}
\subsection{Classification with Extreme Gradient Boost}
At the third level of hierarchy, the discrete latent vector representation of the deep autoencoder is fed into an Extreme Gradient Boost based classification layer\cite{chen2015xgboost,chen2016xgboost} motivated by \cite{zhang2018converting}. It is a regularized gradient boosted decision tree that performs well on structured problems. Since our EEG-phonological pairwise classification has an internal structure involving individual phonemes and words, it seems to be a reasonable choice of classifier. The classifier receives its input from the latent vectors of the deep autoencoder and is trained in a supervised manner to output the final predicted classes corresponding to the speech imagery. 
\section{Experiments and Results}

\subsection{Dataset}
We evaluate our model on a publicly available dataset, KARA ONE ~\cite{zhao2015classifying}, composed of multimodal data for stimulus-based, imagined and articulated speech state corresponding to 7 phonemic/syllabic ( \textit{/iy/}, \textit{/piy/}, \textit{/tiy/}, \textit{/diy/},  \textit{/uw/}, \textit{/m/}, \textit{/n/} 
) as well as 4 words(\textit{pat}, \textit{pot}, \textit{knew} and \textit{gnaw}). The dataset consists of 14 participants, with each prompt presented 11 times to each individual. Since our intention is to classify the phonological categories from human thoughts, we discard the facial and audio information and only consider the EEG data corresponding to imagined speech. It is noteworthy that given the mixed nature of EEG signals, it is reportedly challenging to attain a pairwise EEG-phoneme mapping\cite{sun2016neural}. In order to explore the problem space, we thus specifically target five binary classification problems addressed in \cite{zhao2015classifying,sun2016neural}, i.e presence/absence of consonants, phonemic nasal, bilabial, high-front vowels and high-back vowels. 

\subsection{Training and hyperparameter selection}
We performed two sets of experiments with the single-trial EEG data. In \texttt{PHASE-ONE}, our goals was to identify the best architectures and hyperparameters for our networks with a reasonable number of runs.
For \texttt{PHASE-ONE}, we randomly shuffled and divided the data (1913 signals from 14 individuals) into train (80\%), development (10\%) and test sets (10\%). In \texttt{PHASE-TWO}, in order to perform a fair comparison with the previous methods reported on the same dataset, we perform a leave-one-subject out cross-validation experiment using the best settings we learn from \texttt{PHASE-ONE}. 

The architectural parameters and hyperparameters listed in Table \ref{hyp} were selected through an exhaustive grid-search based on the validation set of \texttt{PHASE-ONE}. We conducted a series of empirical studies starting from single hidden-layered networks for each of the blocks and, based on the validation accuracy, we increased the depth of each given network and selected the optimal parametric set from all possible combinations of parameters. For the gradient boosting classification, we fixed the maximum depth at 10, number of estimators at 5000, learning rate at 0.1,  regularization coefficient at 0.3, subsample ratio at 0.8, and column-sample/iteration at 0.4. We did not find any notable change of accuracy while varying other hyperparameters while training gradient boost classifier.

\begin{table}
\centering
\scriptsize 
 \caption{Results in accuracy on 10\% test data in the first study}
\label{test_result}
\begin{tabular}{llllll}   
\hline
\textbf{Method}  &$\pm$ \textbf{Bilab}& $\pm$ \textbf{Nasal}& \textbf{C/V }&$\pm$ \textbf{/uw/}&$\pm$ \textbf{/iy/ }\\ \hline
LSTM   & 46.07  & 45.31   & 45.83   & 48.44  & 46.88     \\ 
CNN    &59.16     & 57.20   & 67.88    & 69.56    & 68.60    \\ 
CNN+LSTM      & 62.03  & 60.89 & 70.04 & 72.76 &  63.75 \\ 
Our Mixed         & 78.65  & 74.57  & 87.96 & 83.25 & 77.30   \\ \hline

\end{tabular}
\end{table}
\subsection{Performance analysis and discussion}
To demonstrate the significance of the hierarchical CNN-LSTM-DAE method, we conducted separate experiments with the individual networks in \texttt{PHASE-ONE} of experiments and summarized the results in Table \ref{test_result}
From the average accuracy scores, we observe that the mixed network performs much better than individual blocks which is in agreement with the findings in \cite{zhang2018converting}. A detailed analysis on repeated runs further shows that in most of the cases, LSTM alone does not perform better than chance. CNN, on the other hand, is heavily biased towards the class label which sees more training data corresponding to it. Though the situation improves with combined CNN-LSTM, our analysis clearly shows the necessity of a better encoding scheme to utilize the combined features rather than mere concatenation of the penultimate features of both networks.

\begin{figure}[]
\centering
\includegraphics[width=8cm,height=4cm,keepaspectratio]{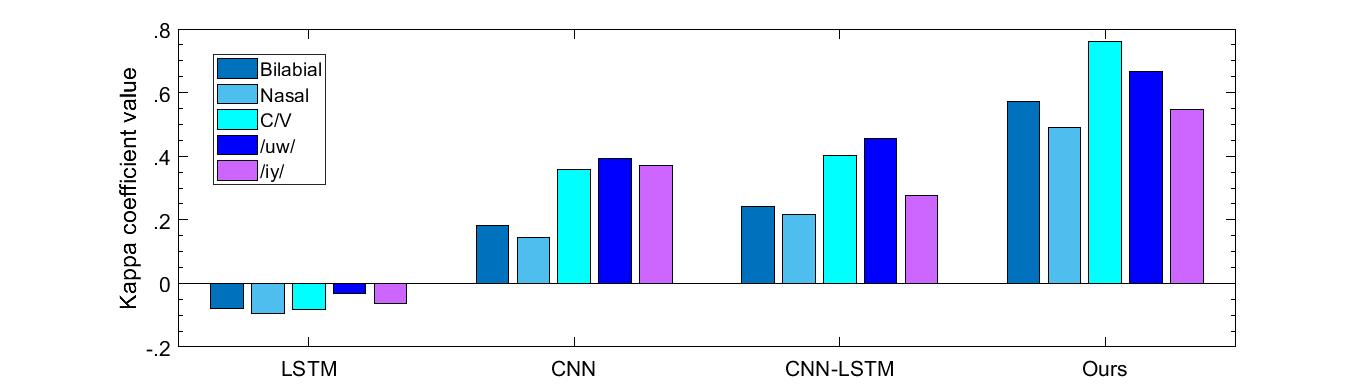}
\caption{Kappa coefficient values for above-chance accuracy based on Table \ref{test_result}} 
\label{fig:kappa}
\end{figure}

The very fact that our combined network improves the classification accuracy by a mean margin of 14.45\% than the CNN-LSTM network indeed reveals that the autoencoder contributes towards filtering out the unrelated and noisy features from the concatenated penultimate feature set. It also proves that the combined supervised and unsupervised neural networks, trained hierarchically, can learn the discriminative manifold better than the individual networks and it is crucial for improving the classification accuracy. 
In addition to accuracy, we also provide the kappa coefficients~\cite{tabar2016novel} of our method in Fig. \ref{fig:kappa}. 
 Here, a higher mean kappa value corresponding to a task implies that the network is able to find better discriminative information from the EEG data beyond random decisions. The maximum above-chance accuracy (75.92\%) is recorded for presence/absence of the vowel task and the minimum (49.14\%) is recorded for the $\pm{nasal}$.

To further investigate the feature representation achieved by our model, we plot T-distributed Stochastic Neighbor Embedding (tSNE) corresponding to $\pm{nasal}$ and V/C classification tasks in Fig. \ref{fig:tSNE} . We particularly select these two tasks as our model exhibits respectively minimum and maximum performance for these two. The tSNE visualization reveals that the second set of features are more easily separable than the first one, thereby giving a rationale for our performance. 

Next, we provide performance comparison of the proposed approach with the baseline methods for \texttt{PHASE-TWO} of our study (cross-validation experiment) in Table \ref{tab:accuracy}. Since the model encounters the unseen data of a new subject for testing, and given the high inter-subject variability of the EEG data, a reduction in the accuracy was expected. However, our network still managed to achieve an improvement of \textbf{18.91},\textbf{ 9.95}, \textbf{67.15}, \textbf{2.83} and \textbf{13.70 \%} over \cite{zhao2015classifying}. Besides, our best model shows more reliability compared to previous works: The standard deviation of our model's classification accuracy across all the tasks is reduced from 22.59\% \cite{zhao2015classifying} and 17.52\%\cite{sun2016neural} to a mere 5.41\%.
\begin{table}[]
\centering
\scriptsize
  \caption{Comparison of classification accuracy}
  \label{tab:accuracy}
  \centering
  \begin{tabular}{lllllll}
    \hline
    &$\pm$ \textbf{Bilabial}& $\pm$\textbf{ Nasal}& \textbf{C/V} &$\pm$ \textbf{/uw/}&$\pm$ \textbf{/iy/}\\\hline
    
    \cite{zhao2015classifying}&56.64&63.5&18.08&79.16&59.6\\ 
    \cite{sun2016neural} &53&47&25&74&53\\
    Ours &\textbf{75.55}& \textbf{73.45}&\textbf{	85.23}&\textbf{	81.99}&	\textbf{73.30}\\\hline
\end{tabular}
\end{table}
\section{Conclusion and future direction}
In an attempt to move a step towards understanding the speech information encoded in brain signals, we developed a novel mixed deep neural network scheme for a number of binary classification tasks from speech imagery EEG data. Unlike previous approaches which mostly deal with subject-dependent classification of EEG into discrete vowel or word labels, this work investigates a subject-invariant mapping of EEG data with different phonological categories, varying widely in terms of underlying articulator motions (eg: involvement or non-involvement of lips and velum, variation of tongue movements etc). Our model takes an advantage of feature extraction capability of CNN, LSTM as well as the deep learning benefit of deep autoencoders. We took \cite{zhao2015classifying,sun2016neural} as the baseline works investigating the same problem and compared our performance with theirs. Our proposed method highly outperforms the existing methods across all the five binary classification tasks by a large average margin of  22.51\%.
\section{Acknowledgments}
This work was funded by the Natural Sciences and Engineering Research Council (NSERC) of Canada and Canadian Institutes for Health Research (CIHR).

\bibliographystyle{IEEEbib}
\bibliography{Saha_ICASSP}

\end{document}